%% file: main.tex
\documentclass[10pt,twocolumn,letterpaper]{article}

\usepackage{cvpr}
\usepackage{times}
\usepackage{epsfig}
\usepackage{graphicx}
\usepackage{amsmath}
\usepackage{amssymb}
\usepackage{bm}
\usepackage{subcaption}
\usepackage{booktabs}
\usepackage{wrapfig,lipsum,booktabs}
\newcommand{\mat}[1]{\mathbf{#1}}
\usepackage[pagebackref=true,breaklinks=true,letterpaper=true,colorlinks,bookmarks=false]{hyperref}

\cvprfinalcopy % *** Uncomment this line for the final submission

\def\cvprPaperID{1990} % *** Enter the CVPR Paper ID here

\newcommand{\eat}[1]{}

\ifcvprfinal\fi
\begin{document}

%%%%%%%%% TITLE
\title{Query-Focused Video Summarization: \\Dataset, Evaluation, and A Memory Network Based Approach}

\author{Aidean Sharghi$^\dagger$, Jacob S.\ Laurel$^\ddagger$\thanks{Jacob S.\ Laurel contributed to this work while he was an NSF REU student at UCF thanks to the support of NSF CNS \#1461121.}, and Boqing Gong$^\dagger$\\
$^\dagger$Center for Research in Computer Vision, University of Central Florida, Orlando, FL 32816\\
$^\ddagger$Department of Computer Science, University of Alabama at Birmingham, AL 35294\\
{\tt\small aidean.sharghi@knights.ucf.edu, jslaurel@uab.edu, bgong@crcv.ucf.edu}
}

\maketitle
%\thispagestyle{empty}

%%%%%%%%% ABSTRACT
\input{abs}

%%%%%%%%% BODY TEXT
\input{intro}
\input{related}

\input{dataset}
\input{approach}

\input{exp_setup}
\input{experiments}
\input{conclude}

{\small
\bibliographystyle{ieee}
\bibliography{egbib}
}

\end{document}

%% file: abs.tex
%!TEX root = main.tex

\begin{abstract}
Recent years have witnessed a resurgence of interest in video summarization. However, one of the main obstacles to the research on video summarization is the user subjectivity --- users have various preferences over the summaries. The subjectiveness causes at least two problems. First, no single video summarizer fits all users unless it interacts with and adapts to the individual users. Second, it is very challenging to evaluate the performance of a video summarizer. 

To tackle the first problem, we explore the recently proposed query-focused video summarization which  introduces user preferences in the form of text queries about the video into the summarization process. We propose a memory network parameterized sequential determinantal point process in order to attend the user query onto different video frames and shots. To address the second challenge, we contend that a good evaluation metric for video summarization should focus on the semantic information that humans can perceive rather than the visual features or temporal overlaps. To this end, we collect dense per-video-shot concept annotations, compile a new dataset, and suggest an efficient  evaluation method defined upon the  concept annotations. We conduct extensive experiments contrasting our video summarizer to existing ones and present detailed analyses about the dataset and the new evaluation method.

\end{abstract}

%% file: intro.tex
% !tex root = 1990.tex

\vspace{-15pt}
\section{Introduction}
\label{sec:intro}

\begin{figure*}[t]
	\centering
	\includegraphics[width=\linewidth]{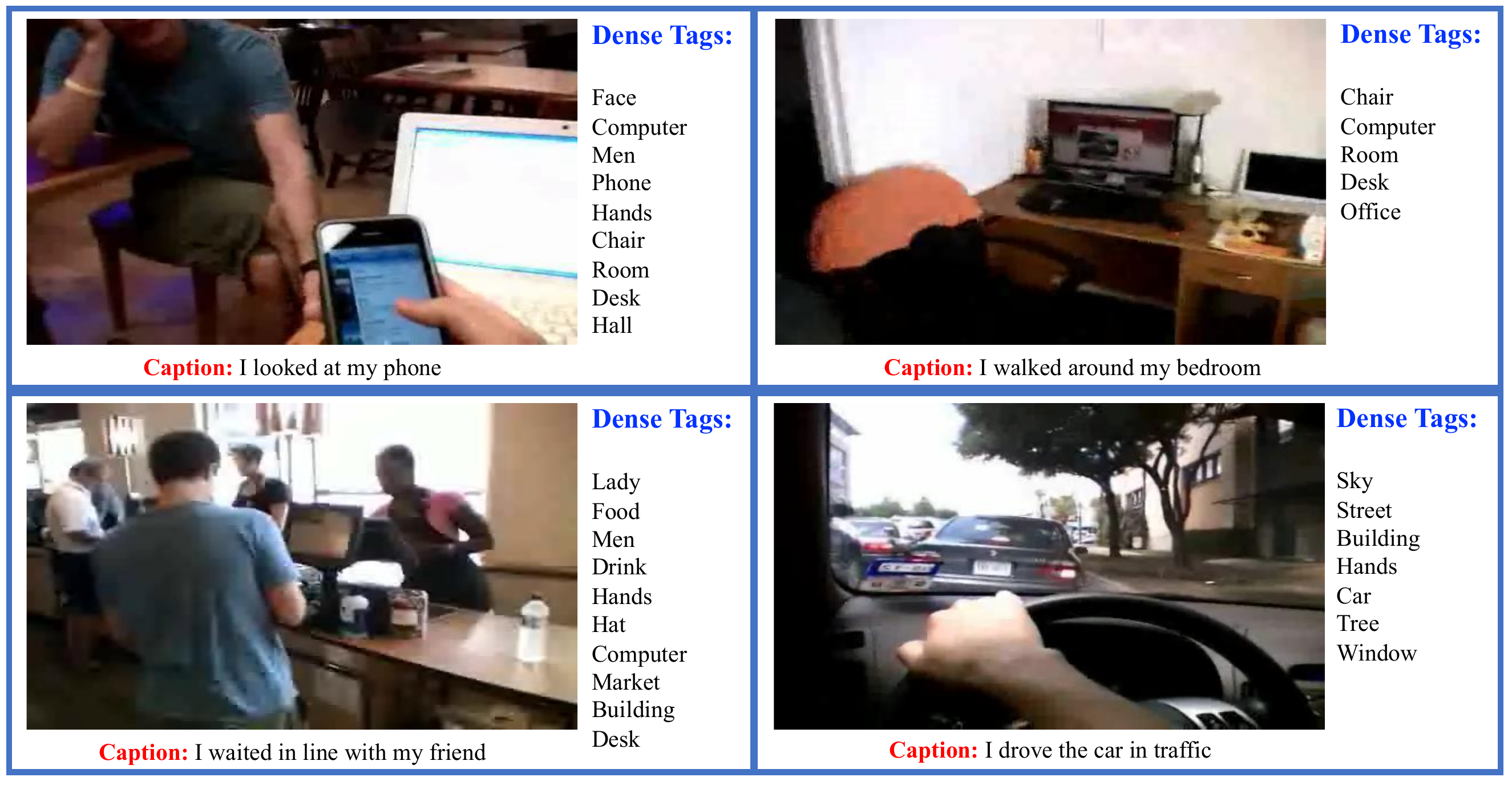}
	\vspace{-18pt}
	\caption{\small{Comparing the semantic information captured by captions in~\cite{yeung2014videoset} and by the concept tags we collected.}}
	\label{fig:captionvstags} 
	\vspace{-10pt}
\end{figure*}

Recent years have witnessed a resurgence of interest in video summarization, probably due to the overwhelming video volumes showing up in our daily life. Indeed,  both consumers and professionals have the access to ubiquitous video acquisition devices nowadays. While the video data is a great asset for information extraction and knowledge discovery, due to its size and variability, it is extremely hard for users to monitor or find the occurrences in it. 

Intelligent video summarization algorithms allow us to quickly browse a lengthy video by capturing the essence and removing redundant information. Early video summarization methods were built mainly upon basic visual qualities (e.g., low-level appearance and motion features)~\cite{goldman2006schematic,gygli2015video,laganiere2008video,liu2002optimization,rav2006making,wolf1996key,zhao2014quasi}, while recently more abstract and higher-level cues are leveraged in the summarization frameworks~\cite{gong2014diverse,khosla2013large,kim2014joint,kwon2012unified,sharghi2016query,xiong2014detecting,yaohighlight,zhang2016summary}.  

However, one of the main obstacles to the research on video summarization is the user subjectivity --- users have various preferences over the summaries they would like to watch. The subjectiveness causes at least two problems. First, no single video summarizer fits all users unless it interacts with and adapts to the users. Second, it is very challenging to evaluate the performance of a video summarizer. 

In an attempt to solve the first problem, we have studied a new video summarization mechanism, query-focused video summarization~\cite{sharghi2016query}, that introduces user preferences in the form of text queries about the video into the summarization process. While this may be a promising direction to \textbf{\em personalize} video summarizers, the experimental study in \cite{sharghi2016query} was conducted on the datasets originally collected for the conventional generic video summarization~\cite{lee2012discovering,yeung2014videoset}. It remains unclear whether the real users would generate distinct summaries for different queries, and if yes, how much the query-focused summaries differ from each other. 

In this paper, we explore more thoroughly the query-focused video summarization and build a new dataset particularly designed for it. While we collect the user annotations, we meet the challenge how to define a good evaluation metric to contrast system generated summaries to user labeled ones --- the second problem above-mentioned due to the user subjectivity about the video summaries. 

%Analogous to query-focused document summarization~\cite{daume2006bayesian,schilder2008fastsum,gupta2007measuring}, 

%For instance, YouTub-ers benefit from this platform in a two-sided fashion, 1) to share their experience with others, 2) learn from other users' experiences. On the other hand, scientists exploit this extensive resource to design various algorithms. 

%The mainstream video summarization approaches are \textit{extractive}; i.e., they generate the summaries by choosing key frames/shots from videos. 
%The integration of two characteristics is crucial in summarization algorithms: 1) {individual importance} of each selected key unit (frame or shot), and 2) combined {diversity} in the summary. %The qualities measured in the video determines how these two characteristics are implemented, resulting in various approaches.  

%In ~\cite{liu2010hierarchical} Liu et al. summarizes the video using frame-level user labeling (relevant/irrelevant) in order to infer objects of interest, which is then used to acquire key units. 

We contend that the pursuit of new algorithms for video summarization has actually left one of the basic problems underexplored, i.e., how to benchmark different video summarizers. User study~\cite{lee2015predicting,lu2013story} is too time-consuming to compare different approaches and their variations at large scale. In the prior arts of automating the evaluation procedure, on one end, a system generated summary has to consist of exactly the same key units (frame or shot)  as in the user summaries in order to be counted as a good one~\cite{chu2015video,song2015tvsum,xu2015gaze}. On the other end, pixels and low-level features are used to compare the system and user summaries~\cite{gong2014diverse,khosla2013large,kim2014joint,zhang2016summary,zhao2014quasi}, whereas it is unclear what features and distance metrics match users' criteria. Some works strive to find a balance between the two extremes, e.g., using the temporal overlap between two summaries to define the evaluation metrics~\cite{gygli2014creating,gygli2015video,potapov2014category,zhang2016video}. However, all such metrics are derived from either the temporal or visual representations of the videos, without explicitly encoding how humans perceive the information --- after all, the system generated summaries are meant to deliver similar information to the users as those directly labeled by the users.

In terms of defining a better measure that closely tracks what humans can perceive from the video summaries, 
we share the same opinion as Yeung et al.'s~\cite{yeung2014videoset}: it is key to evaluate how well a system summary is able to retain the semantic information, as opposed to the visual quantities, of the user supplied video summaries. Arguably, the semantic information is best expressed by the concepts  that represent the fundamental characteristics of what we see in the video at multiple grains, with the focus on different areas, and from a variety of perspectives (e.g., objects, places, people, actions, and their finer-grained entities, etc.). 

Therefore, as our first contribution, we collect dense per-video-shot concept annotations for our dataset. In other words, we represent the semantic information in each video shot by a binary semantic vector, in which the 1's indicate the presence of corresponding concepts in the shot. We suggest a new evaluation metric for the query-focused (and generic) video summarization based on these semantic vector representations of the video shots\footnote{Both the dataset and the code of the new evaluation metric are publicly available at \url{http://www.aidean-sharghi.com/cvpr2017}.}.

In addition, we propose a memory network~\cite{sukhbaatar2015end} parameterized sequential determinantal point process~\cite{gong2014diverse} for tackling the query-focused video summarization. Unlike the hierarchical model in~\cite{sharghi2016query}, our approach does not rely on the costly user supervision about which queried concept appears in which video shot or any pre-trained concept detectors. Instead, we use the memory network to implicitly attend the user query about the video onto different frames within each shot. Extensive experiments verify the effectiveness of our approach.

%Many researchers have taken advantage of underlying power in recurrent neural networks in processing sequences of variable length to design frameworks that are able to perform \textit{reasoning} and \textit{answering} ~\cite{bahdanau2014neural,sukhbaatar2015end,weston2015towards,weston2014memory,xiong2016dynamic}. By presenting the \textit{query} as the question, we take advantage of this powerful platform to obtain query-dependent features. The features are subsequently fed to the summarizer that decides whether a shot must be included in the summary or not. Similar to~\cite{sharghi2016query}, we model the summarization task with \textit{Determinantal Point Process}~\cite{kulesza2012determinantal}, a powerful tool to model negative correlations.

The rest of the paper is organized as follows. We discuss some related works in Section \ref{sec:related}. Section \ref{sec:dataset} elaborates the process of compiling the dataset, acquiring annotations, as well as a new evaluation metric for video summarization. In section \ref{sec:approach} we describe our novel  query-focused summarization model, followed by detailed experimental setup and quantitative results in Sections \ref{sec:expset}.
%%
%\begin{figure*}[t]
%	\includegraphics[width=\textwidth]{./stats}
%	\caption{Distribution of tags in each video. For clarity purposes, the $\log$ of the frequencies are plotted.}
%	\label{fig:dataset_stats}
%\end{figure*}
\begin{figure*}
	\centering
	\includegraphics[width=\linewidth]{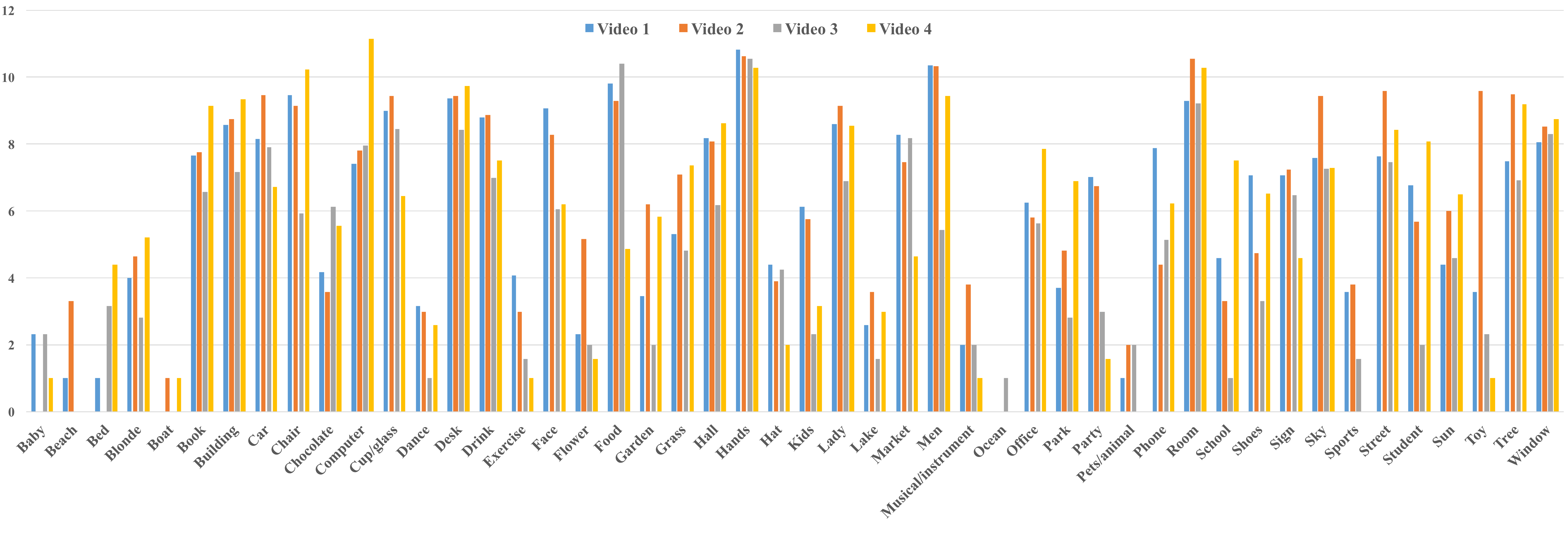}
	\vspace{-25pt}
	\caption{\small{The frequencies of concepts showing up in the video shots, counted for each video separately.}}
	\label{fig:stats} 
	\vspace{-10pt}
\end{figure*}

%% file: related.tex
% !tex root = 1990.tex

\section{Related Work}
\label{sec:related}
We discuss some related works in this section. 

This work extends our previous efforts~\cite{sharghi2016query} on \emph{personalizing} video summarizers.  Both  works explore the query-focused video summarization, but we study this problem more thoroughly in this paper through a new dataset with dense per-video-shot tagging of concepts. Our memory network based video summarizer requires less supervision for training than the hierarchical model in~\cite{sharghi2016query}. 

Unlike our user-annotated semantic vectors for the video shots, Yeung et al.\ asked annotators to caption each video shot using a sentence~\cite{yeung2014videoset}. A single sentence targets only limited information in a video shot and misses many details. Figure~\ref{fig:captionvstags} contrasts the concept annotations in our dataset with the captions for a few video shots. The concept annotations clearly provide a more comprehensive coverage about the semantic information in the shots.

Memory networks~\cite{bahdanau2014neural,sukhbaatar2015end,weston2015towards,weston2014memory,xiong2016dynamic} are versatile in modeling the attention scheme in neural networks. They are widely used to address question answering and visual question answering~\cite{antol2015vqa}. The query focusing in our summarization task is analogous to attending questions to the ``facts'' in the previous works, but  the facts in our context are temporal video sequences. Moreover, we lay a sequential determinantal point process~\cite{gong2014diverse} on top of the memory network in order to promote diversity in the summaries. 

A determinantal point process (DPP)~\cite{kulesza2012determinantal} defines a distribution over the power sets of a ground set that encourages diversity among items of the subsets. There have been growing interest in DPP in machine learning and computer vision~\cite{affandi2014learning,agarwal2014notes,batmanghelich2014diversifying,chao2015large,gartrell2016low,gillenwater2014expectation,DBLP:conf/icml/KuleszaT11,kulesza2011learning,kwok2012priors,li2016fast,mariet2015fixed,mariet2016kronecker,snoek2013determinantal}. Our model in this paper extends DPPs' modeling capabilities through the memory neural network.

%% file: dataset.tex
\section{Dataset}
\label{sec:dataset}

In this section, we provide the details on compiling a comprehensive dataset for video summarization. We opt to build upon the currently existing UT Egocentric (UTE) dataset~\cite{lee2012discovering} mainly for two reasons: 1) the videos are consumer grade, captured in uncontrolled everyday scenarios, and 2) each video is 3--5 hours long and contains a diverse set of events, making video summarization a naturally desirable yet challenging task. In what follows, we first explain how we define a dictionary of concepts and determine the best queries over all possibilities for the query-focused video summarization. Then we describe the procedure of gathering user summaries for the queries. We also show informative statistics about the collected dataset. %We partition the videos to 5 sec long shots equally.

\begin{figure*}
	\centering
	\includegraphics[width=\linewidth]{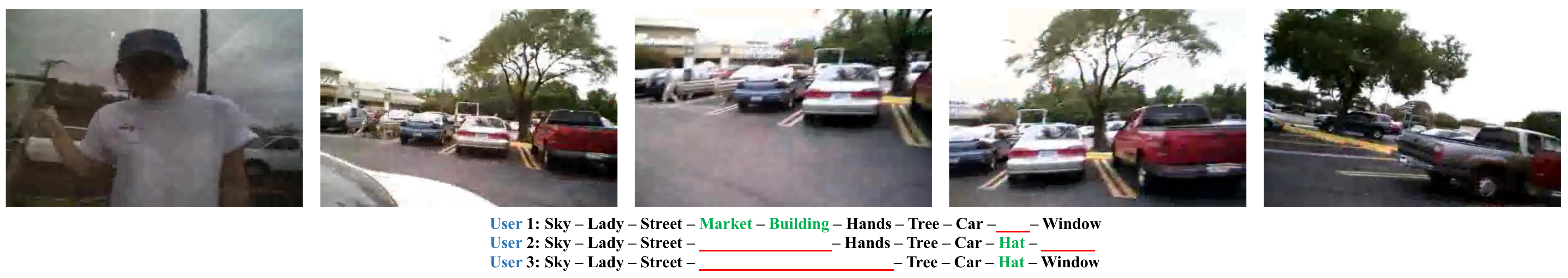}
	\vspace{-18pt}
	\caption{\small{All annotators agree with each other on the prominent concepts in the video shot, while they miss different subtle concepts. }} 	\label{fig:tags} 
	\vspace{-2pt}
\end{figure*}

\subsection{Concept Dictionary and Queries}
\label{subsec:Dict}
We plan to have annotators to transform the semantic information in each video shot to a binary semantic vector (cf.\ Figures~\ref{fig:captionvstags} and \ref{fig:tags}), with 1's indicating the presence of the corresponding concepts and 0's the absence. Such annotations serve as the foundation for an efficient and automatic evaluation method for video summarization described in Section~\ref{subsec:tag}. The key is thus to have a dictionary that covers a wide range and multiple levels of concepts, in order to have the right basis to encode the semantic information. 

\eat{
\begin{table}[t]
	\centering
	\begin{small}
		\begin{tabular}{ll}
			\hline \toprule
			\multicolumn{1}{c}{\normalsize{Concepts}} \\
			\hline 
			Baby, ~Beach, ~Bed, ~Blonde, ~Boat, ~Book, ~Building, ~Car, Chair\\ 
			Chocolate, ~Computer, Cup/glass, ~Dance, Desk, Drink, Exercise\\
			Face, ~Flower, ~Food, ~Garden, ~Grass, ~Hall, ~Hands, ~~Hat, ~Kids\\ 
			Lady, ~Lake, ~Market, ~Men, ~Musical/instrument, ~Ocean, ~Office\\
			Park, ~Party, ~Pets/animal, ~Phone, ~Room, ~School, ~Shoes, ~Sign\\
			Sky, ~Sports, ~Street, ~Student, ~Sun, ~Toy, ~Tree, ~Window\\
			\bottomrule
		\end{tabular}
	\end{small}
	\caption{\small{Dictionary of concepts}} 
	\label{table:dict}
	\vspace{-10pt}
\end{table}
}

In~\cite{sharghi2016query}, we have constructed a lexicon of concepts by overlapping nouns in the video shot captions~\cite{yeung2014videoset} with those in the SentiBank~\cite{borth2013sentibank}. Those nouns serve as a great starting point for us since they are mostly entry-level~\cite{ordonez2013large} words. We prune out the concepts that are weakly related to visual content (e.g., \textsc{''Area''}, which could be interpreted in various ways and applicable to most situations). Additionally, we merge the redundant concepts such as \textsc{''Children''} and \textsc{''Kids''}. We also add some new concepts in order to construct an expressive and comprehensive dictionary. Two strategies are employed to find the new concept candidates. First, after watching the videos, we manually add the concepts that appear for a significant frequency, e.g., \textsc{''Computer''}. Second, we use the publicly available statistics about YouTube and Vine search terms to add the terms that are frequently searched by users, e.g., \textsc{''Pet/Animal''}. The final lexicon is a concise and diverse set of 48 concepts (cf.\ Figure~\ref{fig:stats}) that are deemed to be comprehensive for the UTE videos of daily lives. %The full list of concepts is expanded in . 

\begin{figure*}[t]
	\centering
	\includegraphics[width=\linewidth]{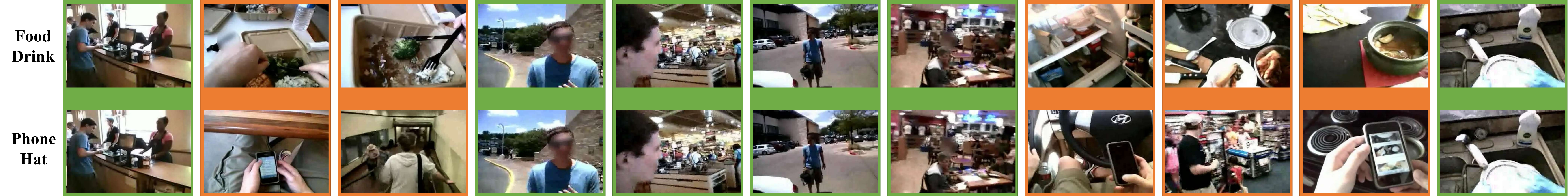}
		\vspace{-18pt} 
	\caption{\small{Two summaries generated by the same user for the queries $\{\textsc{Hat},\textsc{Phone}\}$ and $\{\textsc{Food},\textsc{Drink}\}$, respectively. The shots in the two summaries beside the green bars exactly match each others, while the orange bars show the query-specific shots.}}
	\label{fig:sumcompare}
\vspace{-10pt}
\end{figure*}

We also construct queries, to acquire query-focused user summaries, using two or three concepts as opposed to singletons. Imagine a use case of video search engines. The queries entered by users are often more than one word. For each video, we formalize 46 queries. They cover the following four distinct scenarios: i) all the concepts in the query appear in the same video shots together (15 such queries); ii) all concepts appear in the video but never jointly in a single shot (15 queries), iii) only one of the concepts constituting the query appears in some shots of the video (15 queries), and iv) none of the concepts in the query are present in the video (1 such query). We describe in the Suppl.\ Materials how we obtain the 46 queries to cover the four scenarios. Such queries and their user annotated summaries challenge an intelligent video summarizer from different aspects and extents.

\subsection{Collecting User Annotations}
\label{subsec:GS}

%The ultimate goal in automatic query-focused video summarization is to extract a summary that integrates two components; 1) a collectively diverse and important set of frames/shots that match the query criteria, and 2) a collection of shots/frames in the video that are important in the context of the video. The union of these two components gives a summary that preserves both \textit{\textbf{query}} and the \textit{\textbf{story}}. 

%Given the dictionary size of 48 concepts, it is neither practical nor wise to gather groundtruth summaries for every possible pair.  In order to address this problem, we define a two-stage user annotation task; 1) tagging frames/shots with the concepts in the dictionary, and 2) use tagged shots statistics in order to choose top concept pairs. 

We plan to build a video summarization dataset that offers 1) efficient and automatic evaluation metrics and 2) user summaries in response to different queries about the videos. For the former 1), we collect user annotations about the presence/absence of concepts in each video shot. This is a quite daunting task conditioning on the lengths of the videos and the size of our concept dictionary. We use Amazon Mechanical Turk (MTurk) (\url{http://www.mturk.com/}) for economy and efficiency considerations. For the latter 2), we hire three student volunteers to have better quality control over the labeled video summaries. We uniformly partition the videos to 5-second-long shots.

\vspace{-5pt}
\subsubsection{Shot Tagging: Visual Content to Semantic Vector}
\label{subsec:tag}
\vspace{-5pt}
We ask MTurkers to tag each video shot with all the concepts that are present in it. To save the workers' time from watching the shots, we uniformly extract five frames from each shot. A concept is assumed relevant to the shot as long as it is found in any of the five frames. Figure ~\ref{fig:tags} illustrates the tagging results for the same shot by three different workers. While all the workers captured the prominent concepts like \textsc{Sky}, \textsc{Lady}, \textsc{Street}, \textsc{Tree}, and \textsc{Car},  they missed different subtle ones. The union of all their annotations, however, provides a more comprehensive semantic description about the video shot than that of any individual annotator. Hence, we ask three workers to annotate each shot and take their union to obtain the final semantic vector for the shot. On average, we have acquired $4.13$, $3.95$, $3.18$, and $3.62$ concepts per shot for the four UTE videos, respectively. In sharp contrast, the automatically derived concepts~\cite{sharghi2016query} from the shot captions~\cite{yeung2014videoset} are far from enough; on average, there are only $0.29$, $0.58$, $0.23$, and $0.26$ concepts respectively associated with each shot of the four videos.
\vspace{-13pt}
 %Additionally, we define 46 queries using the concepts for each video and then ask student volunteers to provide the query-focused, generic, and length-budgeted video summaries. The queries and how we select them for the four scenarios presented in Section~\ref{subsec:Dict} are included in the Suppl.\ Materials.

\paragraph{Evaluating video summaries.} Thanks to the dense concept annotations per video shot, we can conveniently contrast a system generated video summary to user summaries according to the semantic information they entail. We first define a similarity function between any two video shots by intersection-over-union (IOU) of their corresponding concepts. For instance, if one shot is tagged by \{\textsc{Car}, \textsc{Street}\} and another by \{\textsc{Street}, \textsc{Tree}, \textsc{Sign}\}, then the IOU similarity between them is ${1}/{4} = 0.25$. 

To find the match between two summaries, it is convenient to execute it by the maximum weight matching of a bipartite graph, where the summaries are on opposite sides of the graph. The number of matched pairs thus enables us to compute precision, recall, and F1 score. Although this procedure has been used in the previous work~\cite{khosla2013large,de2011vsumm}, there the edge weights are calculated by low-level visual features which by no means match the semantic information humans obtain from the videos. In sharp contrast, we use the IOU similarities defined directly over the user annotated semantic vectors as the edge weights.

%Various approaches for automatic summary evaluation exist. Least subjective of all, Yeung et al.~\cite{yeung2014videoset} proposed to evaluate summaries based on the semantic information they carry using ROUGE~\cite{lin2004rouge} metrics. However, as observed in~\cite{chen2015microsoft}, metrics such as ROUGE and BLEU~\cite{papineni2002bleu} poorly correlate with human judgment. Agrawal et al. ~\cite{antol2015vqa} observes that when answering a question about an image, captions are insufficient, i.e. in order to answer the question correctly, one needs to obtain deeper image understanding beyond what captions can capture. Similar to ~\cite{pirsiavash2012detecting}, we argue that providing semantic information in form of binary concept occurrence in an image (or in a video shot) conveys more information than a caption, hence designing a novel evaluation metric that facilitates evaluating system summaries against a gold standard. To this end, we take advantage of shot tags, defining a matching score between two shots by intersection-over-union of their corresponding tag sets. For instance, if $\text{Tags}_i$= \{\textsc{Car}, \textsc{Street}\} and $\text{Tags}_j$=\{\textsc{Street}, \textsc{Tree}, \textsc{Sign}\} represent the set of tags for shots $i$ and $j$ respectively, the similarity between these two shots is reported by $\frac{1}{5} = 0.2$. By computing this similarity measure between all elements in two summaries, and applying a maximum bipartite graph matching algorithm, we can evaluate the summaries.

\vspace{-5pt}
\subsubsection{Acquiring User Summaries}
\label{subsec:user_sums}
In addition to the dense per-video-shot concept tagging, we also ask annotators to label query-focused video summaries for the 46 queries described in Section~\ref{subsec:Dict}.

To ensure consistency in the summaries and better quality control over the summarization process, we switch from MTurk to three student volunteers in our university. We meet and train the volunteers in person. They each summarize all four videos by taking queries into account --- an annotator receives 4 (videos) $\times$ 46 (queries) summarization tasks in total. We thus obtain three user summaries for each query-video pair.  %Hence, we are able to have consistent summaries from each user and also measure inter-user agreement conveniently (cf. Suppl.\ Materials.). 
\eat{
\begin{figure*}[t]
	\centering
	\includegraphics[width=\linewidth]{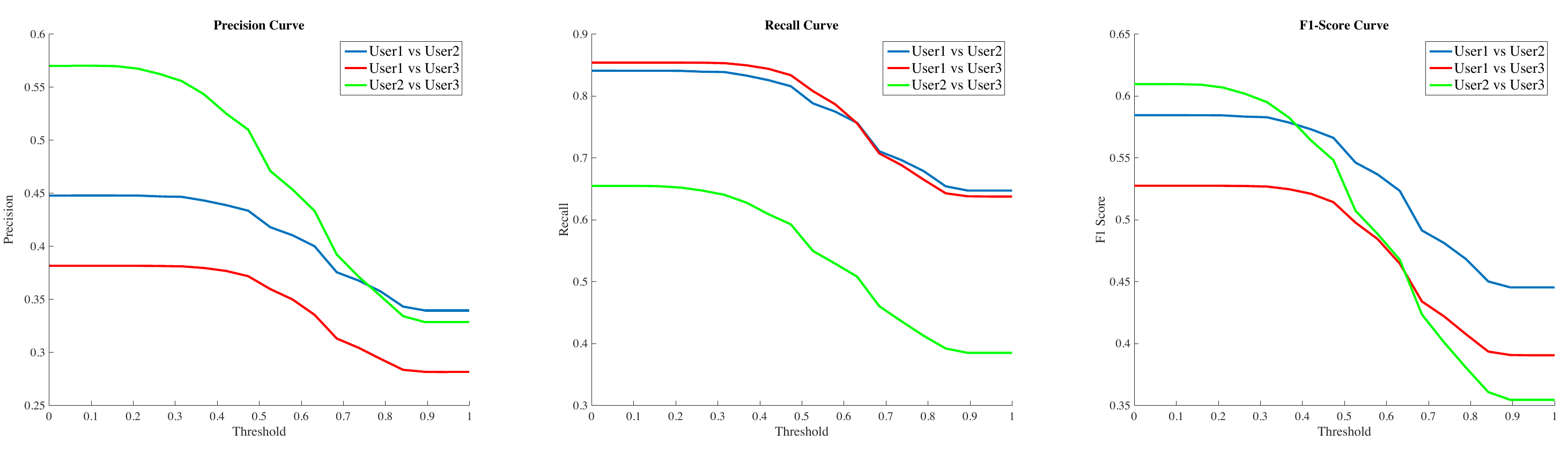}
	\caption{\small{This figure compares and contrasts the summaries labeled by the same user for two queries $\{\textsc{Hat},\textsc{Phone}\}$ and $\{\textsc{Food},\textsc{Drink}\}$, respectively. Pairs that are marked with green margin exactly match in the user summaries while orange margin depicts the query-relevant shots in the summaries.}}
	\label{fig:interuser} 
\end{figure*}
}

However, we acknowledge that it is infeasible to have the annotators to summarize all the query-video pairs from scratch --- the UTE videos are each 3--5 hours long. To overcome this issue, we expand each temporal video to a set of static key frames. First, we uniformly extract five key frames to represent each shot in the same way as in Section~\ref{subsec:tag}. Second, we pool all the shots corresponding to the three textual summaries~\cite{yeung2014videoset} as the initial candidate set. Third, for each query, we further include all the shots that are relevant to it into the set. A shot is relevant to the query if the intersection of the concepts  associated with it and the query is nonempty. As a result, we have a set of candidate shots for each query that covers the main story in the video as well as those of relevance to the query. The annotators summarize the video by removing redundant shots from the set. There are $2500$ to $3600$ shots in the candidate sets, and the summaries labeled by the participants contain only $71$ shots on average. 
\vspace{-12pt}

\begin{table}[t]\centering
	\small
		\caption{\small{Inter-user agreement evaluated by F1 score (\%) (U1, U2, and U3: the three student volunteers,  O: the oracle summary).}} 	\label{tab:interuser} 
		\vspace{-10pt}
	\begin{tabular}{cccccccccccccccccccccc}\toprule
		U1-U2 & U1-U3 & U2-U3 & U1-O & U2-O & U3-O
		\\ \midrule
		%Vid1 & 58.4 & 52.7 & 61.0 & 66.2 & 81.7 & 75.1\\
		%Vid2 & 54.9 & 62.0 & 60.3 & 69.2 & 75.5 & 82.3\\
		%Vid3 & 59.1 & 64.9 & 69.9 & 69.7 & 81.4 & 86.3\\ 
		%Vid4 & 48.7 & 43.8 & 59.5 & 54.8 & 80.4 & 76.6\\
		%\bottomrule
		 55.27 & 55.85 & 62.67 & 64.97 & 79.75 & 80.07\\
		\bottomrule
	\end{tabular}
	\label{table:InterUser}
	\vspace{-15pt}
\end{table}

\paragraph{Oracle summaries.} 
Supervised video summarization methods~\cite{gong2014diverse,gygli2015video,sharghi2016query,zhang2016summary,zhang2016video} often learn from one summary per video, or per query-video pair in query-focused summarization, while we have three user generated summaries per query. We aggregate them into one, called the oracle summary, per query-video pair by a greedy algorithm. The algorithm starts from the common shots in the three user summaries. It then greedily chooses one shot every time such that this shot gives rise to the largest marginal gain over the evaluated F1 score. We leave the details to the Suppl.\ Materials. The oracle summaries achieve better agreements with the users than the inter-user consensus (cf.\ Table~\ref{tab:interuser}). 
\vspace{-12pt}

%As mentioned in Section \ref{subsec:user_sums}, we collect 3 user summaries per query-video pair. We adopt a greedy approach proposed in~\cite{kulesza2012determinantal} to merge all 3 summaries of a query-video pair into an \textit{oracle} summary that maximally agrees with them. The greedy algorithm starts with an empty set, and at each iteration, the video shot that gives rise to largest gain in the F1-score (of our proposed metric) is added to the set. The algorithm stops when no gain is achieved in an iteration.

\begin{table}
	\small
			\centering
		\caption{\small{The average lengths and standard deviations of the summaries for different queries. }}
		\vspace{-10pt}
	\label{table:sumstats}
%\ra{1.2}
	\begin{tabular}{ccccccccc}
	\toprule
		& User 1 & User 2 & User 3 & Oracle \\ \cmidrule{2-5}
		Vid1 & \small{143.7$\pm$}32.5 & \small{80.2$\pm$}47.1 & \small{62.6$\pm$}15.7 & \small{82.5$\pm$}33.9 \\%&& 54.7$\pm$0.13 & 63.4$\pm$0.11 & 61.8$\pm$0.08 & 64.3$\pm$0.09\\
		Vid2 & 103.0$\pm$45.0 & 49.9$\pm$25.2 & 64.4$\pm$11.7 & 64.1$\pm$11.7 \\%&& 52.0$\pm$0.13 & 65.6$\pm$0.08 & 59.6$\pm$0.1 & 63.2$\pm$0.08\\
		Vid3 & 97.3$\pm$38.9  & 50.1$\pm$9.6  & 58.4$\pm$9.3 & 59.2$\pm$9.6 \\%&& 56.2$\pm$0.13 & 67.4$\pm$0.06 & 66.7$\pm$0.08 & 68.2$\pm$0.06\\ 
		Vid4 & 79.9$\pm$30.3 & 34.4$\pm$7.3 & 28.9$\pm$8.7 & 35.6$\pm$8.5 \\%&& 42.9$\pm$0.15 & 54.3$\pm$0.08 & 52.4$\pm$0.09 & 54.1$\pm$0.07\\
		\bottomrule
	\end{tabular}
	\vspace{-15pt}
\end{table}

\paragraph{Summaries of the same video differ due to queries.}
Figure~\ref{fig:sumcompare} shows two summaries labeled by the same user for two distinct queries, $\{\textsc{Hat},\textsc{Phone}\}$ and $\{\textsc{Food},\textsc{Drink}\}$. Note that the summaries both track the main events happening in the video while they differ in the query-specific parts. Besides, table~\ref{table:sumstats} reports the means and standard deviations of the lengths of the summaries per video per user. We can see that the queries highly influence the resulting summaries; the large standard deviations attribute to the queries. 
\vspace{-22pt}

%\paragraph{Inter-user agreement.} 

\paragraph{Budgeted summary.}
For all the summaries thus far, we do not impose any constraints over the total number of shots to be included into the summaries. After we receive the annotations, however, we let the same participants further reduce the lengths of their summaries to respectively 20 shots and 10 shots. We call them \emph{budgeted} summaries and leave them for future research. %work to explore their properties and usages. %examining their properties and usages in the future work. 

%According to ~\cite{sharghi2016query}, in query-focused video summarization, an ideal summary includes two key components; 1) contextually important set of shots, and 2) query-relevant shots. For the first component, we use the summaries from , where 3 textual summaries per each video in the dataset is presented. These summaries are subsets of the dense (shot-by-shot) textual description provided for each video, hence, we were able to map back from textual summaries to indices of selected shots. The union of 3 extracted video summaries represent a good candidate for the first component, however, we still need to ask users to remove the redundancy in them. For query-relevant shots (second component), for each query we extract all the shots that are relevant to it; relevant shots for scenarios, i) are the shots that are tagged with both concepts of interest in the query, ii) we combine the shots corresponding to each of the concepts in the query (knowing the intersection of them is empty), iii) since only one of the concepts in the query is present in the video, we include the shots corresponding the the appearing concept, and iv) we only put the shots that were initially identified contextually important (first component). Finally, for each query, the union of query-relevant shots and contextually important shots is released to the users for summarization. Besides, table ~\ref{table:sumstats} indicates that queries can highly influence the resulting summary by reporting the mean and standard deviation of the obtained summaries.

%% file: approach.tex
% !tex root = 1990.tex

\section{Approach}
\label{sec:approach}
We elaborate our approach to the query-focused video summarization in this section. Denote by ${\cal V} = \{{\cal V}_t\}_{t=1}^T$ a video that is partitioned to $T$ segments, and by $q$ the query about the video. In our experiments, every segment ${\cal V}_t$ consists of $10$ video shots each of which is 5-second long and is used in Section~\ref{subsec:GS} to collect the concept annotations. 

\begin{figure*}[t]
	\includegraphics[width=\textwidth]{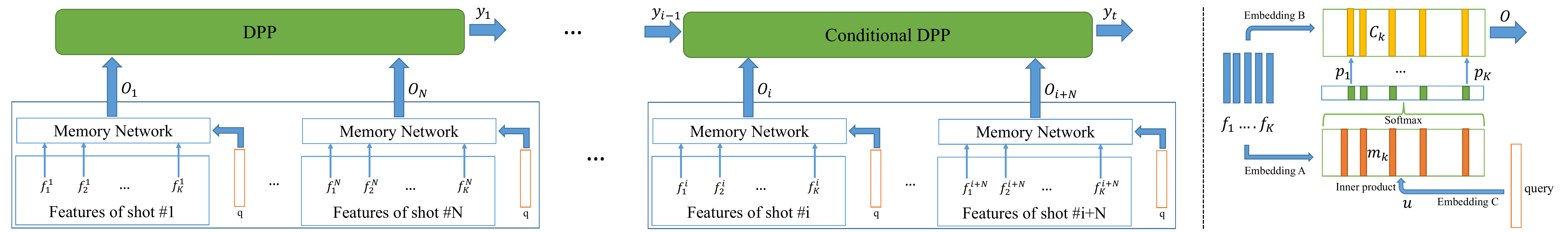}
	\caption{\small{Our query-focused video summarizer: Memory network (right) parameterized sequential determinantal point process (left).}}
	\label{fig:framework}
	\vspace{-10pt} 
\end{figure*}

\subsection{Query Conditioned Sequential DPP}
The sequential determinantal point process (DPP)~\cite{gong2014diverse} is among the state-of-the-art models for generic video summarization. We condition it on the query $q$ as our overarching video summarization model,
\begin{align}
&P(Y_1=\bm{y}_1, Y_2 = \bm{y}_2, \cdots, Y_T=\bm{y}_T | \mathcal{V}, q) \\ 
= &P(Y_1=\bm{y}_1 | \mathcal{V}_1, q)\prod_{t=2}^T P(Y_t = \bm{y}_t | \mathcal{V}_t, \bm{y}_{t-1}, q)
\end{align}
where the $t$-th DPP variable $Y_t$ selects subsets from the $t$-th segment ${\cal V}_t$, i.e., $\bm{y}_t\subseteq{\cal V}_t$, and the distribution $P(Y_t = \bm{y}_t | \mathcal{V}_t, \bm{y}_{t-1}, q)$ is specified by a conditional DPP~\cite{kulesza2012determinantal},
\begin{align}
P(Y_t = \bm{y}_t | \mathcal{V}_t, \bm{y}_{t-1}, q) = \frac{\det [\bm{L}(q)]_{\bm{y}_t\cup\bm{y}_{t-1}}}{\det\big(\bm{L}(q) + \bm{I}_t\big)}.
\end{align}
The nominator on the right-hand side is the principle minor of the (L-ensemble) kernel matrix $\bm{L}(q)$ indexed by the subsets $\bm{y}_t\cup\bm{y}_{t-1}$. The denominator calculates the determinant of the sum of the kernel matrix and a corrupted identity matrix whose elements indexed by $\bm{y}_{t-1}$ are 0's. Readers are referred to the great tutorial~\cite{kulesza2012determinantal} on DPP for more details.  

Note that the DPP kernel $\bm{L}(q)$ is parameterized by the query $q$. We have to carefully devise the way of parameterizing it in order to take account of the following properties. In query-focused video summarization,  a user selects a shot to the summary for two possible reasons. One is that the shot is quite related to the query and thus becomes appealing to the user. The other may attribute to the contextual importance of the shot; e.g., the user would probably  choose a shot to represent a prominent event in the video even if the event is not quite relevant to the query. To this end, we use a memory network to model the two types of importance (query-related and contextual) of a video shot simultaneously.

\subsection{Memory Network to Parameterize DPP Kernels}
The memory network~\cite{sukhbaatar2015end} offers a neural network architecture to naturally attend a question to ``facts'' (cf.\ the rightmost panel of Figure~\ref{fig:framework}). In our work, we shall measure the relevance between the query $q$ and a video shot and incorporate such information into the DPP kernel $\bm{L}(q)$. Therefore, it is straightforward to substitute the question in memory network by our query, but the ``facts'' are less obvious. 

As discussed in Section~\ref{subsec:Dict}, there could be various scenarios for a query and a shot. All the query concepts may appear in the shot but possibly in different frames; one or two concepts of the query may not be present in the shot; it is also possible that none of the concepts are relevant to any frame in the shot. In other words, the memory network is supposed to screen all the video frames in order to determine  the shot's relevance to the query. Hence, we uniformly sample 8 frames from each shot as the ``facts''.  The video frames are represented using the same feature as~\cite{sharghi2016query} (cf.\ $\bm{f}_1,\cdots,\bm{f}_K$ on the rightmost panel of Figure~\ref{fig:framework}).

The memory network takes as input the video frames $\{\bm{f}_k\}$ of a shot and a query $q$. The frames are transformed to memory vectors $\{\bm{m}_k\}$ through an embedding matrix $A$. Similarly, the query ${q}$, represented by a binary indication vector, is mapped to the internal state $\bm{u}$ using an embedding matrix $C$. The attention scheme is implemented simply by a dot product followed by a softmax function,
\begin{equation}
p_k = \text{Softmax}(\bm{u}^T\bm{m}_k),
\end{equation}
where $p_k$ carries how much attention the query $q$ incurred over the frame $\bm{f}_k$. 

Equipped with the attention scores $\{p_k\}$, we assemble another embedding $\{\bm{c}_k\}$ of the frames, obtained by the mapping matrix $B$ in figure~\ref{fig:framework}, into the video shot representation  $\bm{o}$:
\begin{equation}
\bm{o} = \sum_{k}p_i\bm{c}_k,
\end{equation}
which is conditioned on the query $q$ and entails the relevance strength of the shot to the query. As a result, we expect the DPP kernel parameterized by the following
\begin{equation}
[\bm{L}(q)]_{ij} = \bm{o}_i^T {D}^T {D} \bm{o}_j \label{eWeightCombine}
\end{equation}
is also flexible in modeling the importance of the shots to be selected into the video summary. Here $i$ and $j$ index two shots, and ${D}$ is another embedding matrix. Note that the contextual importance of a shot can be inferred from a shot's similarities to the others by the kernel matrix, while the query-related importance is mainly by the attention scheme in the memory network.

%In ~\cite{trenn2008multilayer}, it is proven that for high approximation orders, one needs two fully connected layers in the network, thus, having embedding matrices \textbf{B} and \textbf{D} in a consecutive order in the network, helps improving the overall performance.

\subsection{Learning and Inference}
We learn the overall video summarizer, including the sequential DPP and the memory network, by maximizing the log-likelihood of the user summaries in the training set. We use stochastic gradient descent with mini-batching to optimize the embedding matrices $\{A, B, C, D\}$. The learning rates and numbers of epochs are chosen using the validation set. At the test stage, we sequentially visit the video segments ${\cal V}_1, \cdots, {\cal V}_T$ and select shots from them using the learned summarization model.

It is notable that our approach requires less user annotations than the SH-DPP~\cite{sharghi2016query}. It learns directly from the user summaries and implicitly attend the queries to the video shots. However, SH-DPP requires very costly annotations about the relevances between video shots and queries. Our new dataset does supply such supervisions, so we shall include SH-DPP as the baseline method in our experiments.

%% file: exp_setup.tex
% !tex root = 1990.tex

\section{Experimental Results}
\label{sec:expset}
\begin{table*}[t]
	\centering
	\small
	\caption{\small{Comparison results for query-focused video summarization (\%). }}
	\label{table:results}
	\vspace{-10pt}
	\begin{tabular}{@{}rrrrcrrrcrrr@{}}\toprule
		& \multicolumn{3}{c}{SeqDPP~\cite{gong2014diverse}} & \phantom{abc}& \multicolumn{3}{c}{SH-DPP~\cite{sharghi2016query}} &
		\phantom{abc} & \multicolumn{3}{c}{\textbf{Ours}}\\ \cmidrule{2-4} \cmidrule{6-8} \cmidrule{10-12}
		& Precision & Recall & F1 && Precision & Recall & F1 && Precision & Recall & F1\\ \midrule
		
		Vid1 & \textbf{53.43} & 29.81 & 36.59 && 50.56 & 29.64 & 35.67 && 49.86 & \textbf{53.38} & \textbf{48.68}\\
		Vid2 & \textbf{44.05}& 46.65 & \textbf{43.67} && 42.13& 46.81& 42.72&& 33.71& \textbf{62.09}& 41.66\\
		Vid3 & 49.25& 17.44 & 25.26 && 51.92& 29.24& 36.51&& \textbf{55.16} & \textbf{62.40} & \textbf{56.47} \\ 
		Vid4 & 11.14 & \textbf{63.49} & 18.15 && 11.51& 62.88& 18.62&& \textbf{21.39}& 63.12& \textbf{29.96}\\
		\bottomrule
		Avg. & 39.47 & 39.35 & 30.92 && 39.03 & 42.14 & 33.38 && \textbf{40.03} & \textbf{60.25} & \textbf{44.19} \\
		\bottomrule
	\end{tabular}
	\vspace{-2pt}
\end{table*}
\eat{
\begin{table*}[t]\centering
	\small
	\caption{\small{Testing effectiveness of individual components in our proposed attention-based query-focused summarizer.}}
	\label{table:elementtest}
	\vspace{-10pt} 
	\begin{tabular}{@{}rrrrcrrrcrrr@{}}\toprule
		& \multicolumn{3}{c}{NoAttention} & \phantom{}& \multicolumn{3}{c}{-Emb $D$} &
		\phantom{} & \multicolumn{3}{c}{EmbSize 256}\\ \cmidrule{2-4} \cmidrule{6-8} \cmidrule{10-12}
		& Precision & Recall & F1 && Precision & Recall & F1 && Precision & Recall & F1\\ \midrule
		
		Vid1 & 39.68 & 61.21 & 45.66 && 34.58 & 66.28 & 42.95 && 44.12 & 59.44 & 47.64\\
		Vid2 & 25.24 & 67.04 & 35.38 && 24.40 & 68.78 & 34.31 && 30.68 & 65.53 & 39.91\\
		Vid3 & 36.67 & 74.27 & 47.51 && 51.26 & 64.60 & 54.46 && 53.88 & 62.58 & 55.61 \\ 
		Vid4 & 14.32 & 65.58 & 22.38 && 15.36 & 67.31 & 23.52 && 17.01 & 66.44 & 25.61\\
		\bottomrule
		Avg. & 28.98 & 67.03 & 37.73 && 31.4 & 66.74 & 38.81 && 36.42 & 63.5 & 42.19\\
		\bottomrule
	\end{tabular}
\end{table*}
}
We report experimental setup and results in this section.
\vspace{-10pt}
\paragraph{Features.} We extract the same type of features as used in the existing SH-DPP method~\cite{sharghi2016query} in order to have fair comparisons. First, we employ 70 concept detectors from SentiBank~\cite{borth2013sentibank} and use the detection scores for the features of each key frame (8 key frames per 5-second-long shot). However, it is worth  mentioning that our approach is not limited to using concept detection scores and, more importantly unlike SH-DPP, does not rely on the per-shot annotations about the relevance to the query --- the per shot user labeled semantic vectors serve for evaluation purpose only. Additionally, we extract a six dimensional contextual feature vector per shot as the mean-correlations of low-level features (including color histogram, GIST~\cite{oliva2001modeling}, LBP~\cite{ojala2002multiresolution}, Bag-of-Words, as well as an attribute feature~\cite{yu2013designing}) in a temporal window whose size varies from 5 to 15 shots. The six-dimensional contextual features are appended to the key frame features in our experiments.

%Due to usage of concept detection scores, including the contextual features is crucial to the model, more specifically to enforce diversity. On the other hand, the concept scores are important to find shots that much query criteria.
\vspace{-10pt}
\paragraph{Data split.} We run four rounds of experiments each leaving one video out for testing and one for validation, while keeping the remaining two for training. Since our video summarizer and the baselines are sequential models, the small number (i.e., two) of training videos is not an issue as the videos are extremely long, providing many variations and supervisions at the training stage. 

%In query-focused video summarization, it is essential to have a feature that incorporate both low-level and high-level features. Low-level components are mandatory due to the fact that they can be used to represent very basic temporal information, showing how similar a shot is compared to its neighboring shots. It is the low-level cues that helps summarization methods to remove redundancy in the summaries; i.e. if a shot is included in the summary, no visually similar and temporally close shot to it should be included as well. On the other hand, high-level components of the feature helps the framework to identify matching between the user query and the shot, making query-relevant summarization feasible.

%We opt to use the same features as ~\cite{sharghi2016query}, as it incorporates both feature components required for query-focused video summarization. The features are 76 dimensional, where first 6 components represent the mean-correlation over a set of low-level features including color histogram, GIST, LBP, Bag-of-Words, as well as an attribute feature. By varying the temporal window size from 5 to 15 and computing mean correlation between the features falling in the temporal window range, we acquire the first 6 components. The remaining 70 components are also the same concept scores from SentiBank~\cite{borth2013sentibank} detectors.

\subsection{Comparison Results}

\begin{table*}[t]\centering
	\small
\caption{\small{Comparison results for generic video summarization, i.e., when no video shots are relevant to the query.}}
	\label{table:glob}
	\vspace{-10pt}
	\begin{tabular}{@{}rrrrcrrrcrrr@{}}\toprule
		& \multicolumn{3}{c}{SubMod~\cite{gygli2015video}} & \phantom{abc}& \multicolumn{3}{c}{Quasi~\cite{zhao2014quasi}} &
		\phantom{abc} & \multicolumn{3}{c}{\textbf{Ours}}\\ \cmidrule{2-4} \cmidrule{6-8} \cmidrule{10-12}
		& Precision & Recall & F1 && Precision & Recall & F1 && Precision & Recall & F1\\ \midrule
		
		Vid1 & 47.86 & 51.28 & 49.51 && 57.37 & 49.36 & 53.06 && \textbf{65.88} & 59.75 & \textbf{62.66}\\
		Vid2 & 56.53 & 46.50 & 51.03 && 46.75& 63.34& \textbf{53.80} && 35.07 & \textbf{67.31} & 46.11\\
		Vid3 & 62.46 & 66.72 & 64.52 && 53.93 & 46.44 & 49.91&& \textbf{65.95} & 53.12 & \textbf{58.85} \\ 
		Vid4 & \textbf{34.49} & 37.25 & \textbf{35.82} && 13.00 & \textbf{77.88} & 22.31 && 22.29 & 67.74 & 33.5\\
		\bottomrule
		Avg. & \textbf{50.34} & 50.44 & 50.22 && 42.76 & 59.25 & 44.77 && 47.3 & \textbf{61.98} & \textbf{50.29}\\
		\bottomrule
	\end{tabular}
\end{table*}

%However, it is wise to train supervised methods using one summary instead of all 3. Hence, we employ the same greedy approach to construct oracle summaries with the slight difference that we use our novel evaluation metric (for more details please refer to the reference ~\cite{kulesza2012determinantal}).

%% file: experiments.tex
% !tex root = 1990.tex
\paragraph{Query-focused video summarization.} We contrast our video summarizer, the memory-network based sequential determinantal point process, to several closely related methods. We first include SH-DPP~\cite{sharghi2016query}, the most recent approach to the query-focused video summarization. Our model improves upon SeqDPP~\cite{gong2014diverse} by taking the query into account and parameterizing the DPP kernel by the memory network. SeqDPP is thus directly comparable to ours. We concatenate the query features (binary indication vectors) with the shot features and input them to SeqDPP and SH-DPP. We set the same dimensionality for all the embedding spaces in our and the two baseline methods. It turns out the 128D embeddings are chosen due to their performances on the validation videos. %, embedding space sizes (matrices $A$,$C$,$B$, and $D$) to $128$. For fair comparison, we set the embedding matrix sizes in SeqDPP and SHDPP to $128$ as well.

\begin{figure}[t]
	\centering
	\vspace{-18pt} 
	\includegraphics[width=\linewidth]{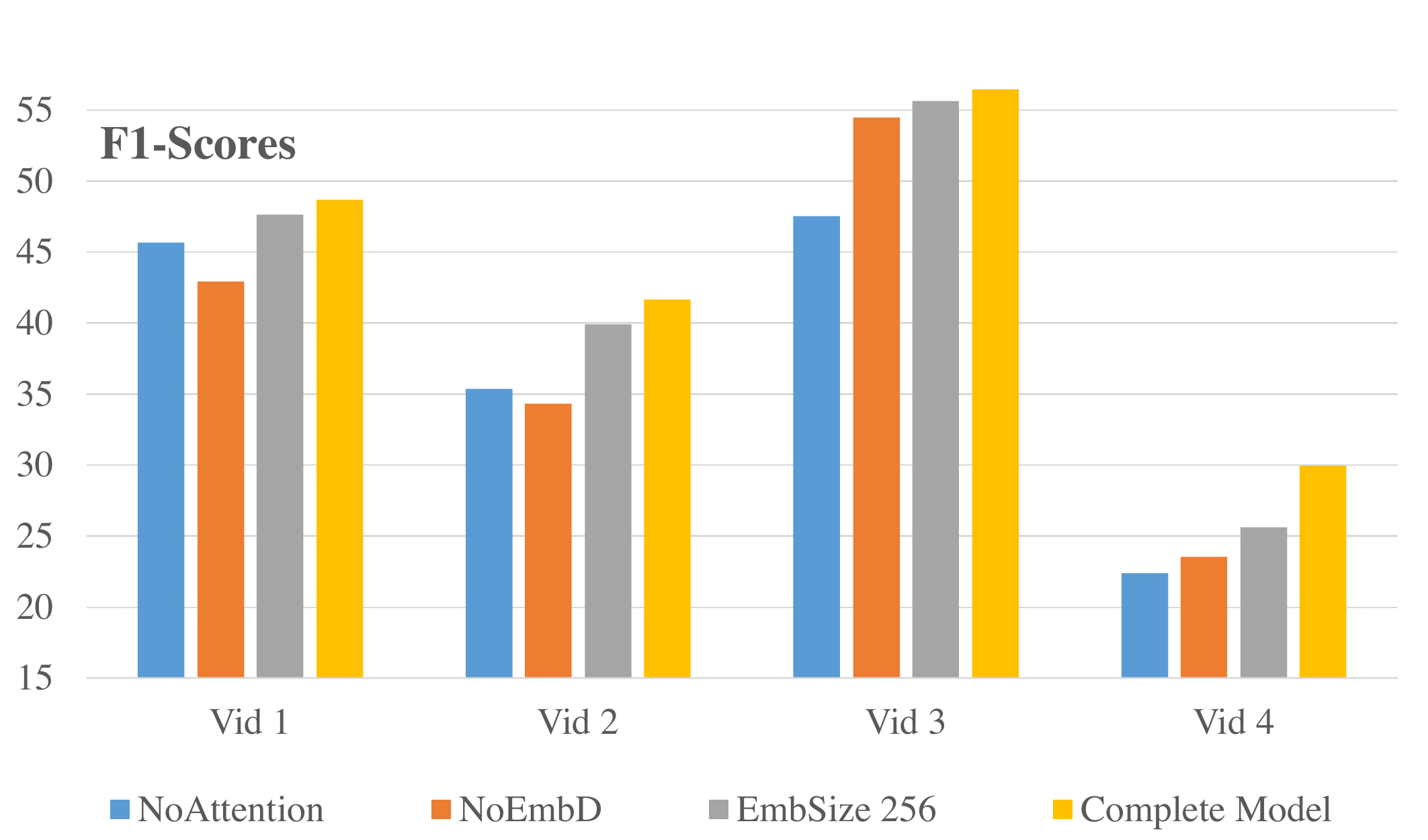}
	\vspace{-18pt}
	\caption{\small{The Effectiveness of various individual components in our proposed video summarizer.}}
	\vspace{-10pt}
	\label{fig:element_test} 
\end{figure}

Table~\ref{table:results} compares the performances of the three video summarizers. Each video is taken in turn as the test video and the corresponding results are shown in each row. The average results are included as the last row. Precision, recall, and F1 score are reported for all the video summarizers. Our approach outperforms the other two by a large margin (more than 10\% F1 score on average). It seems like Video 4 is especially challenging for all the methods. For Video 2, our summarizer generates a little longer summaries than the others do. In the future work, we will explore how to control the summary length in the sequential DPP model.
\vspace{-20pt}
\paragraph{Component-wise analyses.} To investigate how each component in our framework contributes to the final results, we conduct more experiments by either removing or modifying them. Figure~\ref{fig:element_test} shows the corresponding results. 

The main benefit from the memory network is the attention mechanism (cf.\ equation~(\ref{eWeightCombine})). If we instead use a uniform distribution for the attention scores $\{p_i\}$ and append the query information  $\bm{u}$ directly after the memory network output $\bm{o}$, the results become worse on all the four videos (cf.\ NoAttention in Figure~\ref{fig:element_test}). The NoEmbD results are obtained after we remove the last embedding matrix $\bm{D}$ when we compute the DPP kernels. Finally, EmbSize 256 are the results when we change the 128D embeddings in our approach to 256D. The performance drops from our complete model verify that all the corresponding components are complementary, jointly contributing to the final results.

\begin{figure*}[h]
	\centering
	\vspace{-5pt} 
	\includegraphics[width=0.9\linewidth]{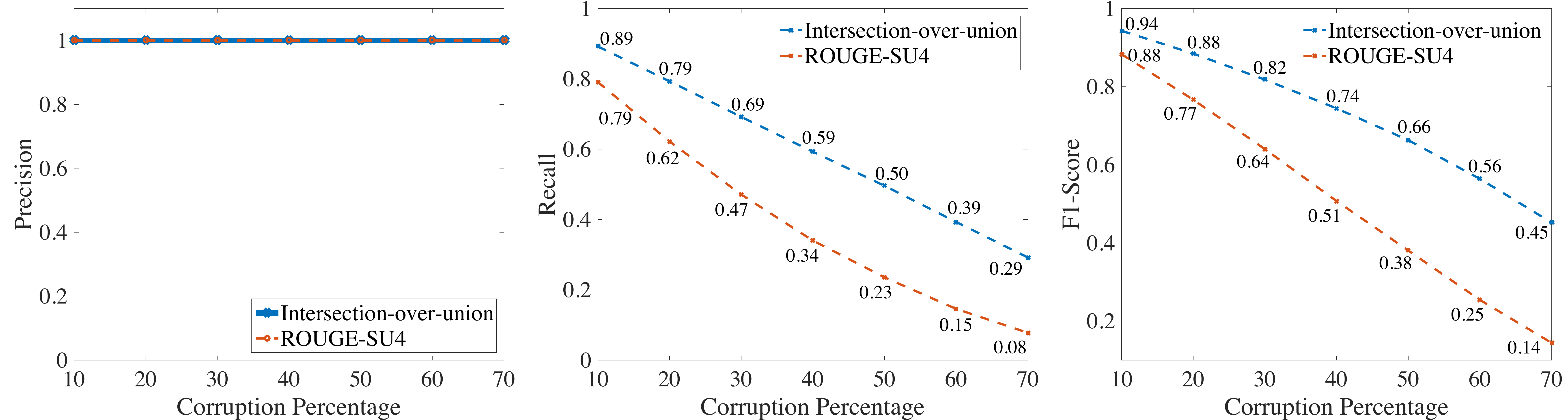}
	\caption{\small{A nice behavior of our evaluation metric. When we randomly remove video shots from the user summaries, the recall between the original user summaries and the corrupted ones decreases almost linearly. The evaluation by ROUGE-SU4~\cite{yeung2014videoset} is included for reference.}}
	\vspace{-10pt}
	\label{fig:Del} 
\end{figure*}
\vspace{-10pt}
\paragraph{Generic video summarization.} Recall that our queries incur four different scenarios (cf.\ Section~\ref{subsec:Dict}). When there are no video shots relevant to the query, it reduces to the generic video summarization in some sort. We single out such queries and contrast our summarizer to some existing and recent methods for generic video summarization: SubMod~\cite{gygli2015video} which employs submodular functions to encourage diversity and Quasi~\cite{zhao2014quasi} which is an unsupervised method based on group sparse coding. Unlike the DPP type of summarizers, the baseline methods here are not able to automatically determine the lengths of the summaries. We tune the threshold parameter in Quasi such that the output lengths are no more or less than the oracle summary by 20 shots. For SubMod we set the budget parameter such that it generates summaries that are exactly as long as the oracle summaries. As shown in Table~\ref{table:glob}, our approach still gives the best overall performance even though we reveal the oracle sumamries' lengths to the baseline methods, probably due to its higher neural network based modeling capacity.

%Aside from the approaches that are capable of generating query-focused summaries, we compare our approach with two methods that are agnostic to this problem, however, they are strong baselines for general (as opposed to query-focused) video summarization.~\cite{zhao2014quasi} is an unsupervised method based on group sparsity coding, and SubMod~\cite{gygli2015video} is a supervised approach that learns a convex combination of various objective functions from user summaries. These approaches cannot be adapted to do query-focused video summarization due to their nature (if the query and the features are initially in the same space, meaning we have high-level concept detectors for each element in the dictionary, they can be adapted to the problem as shown in~\cite{sharghi2016query}), hence, we only evaluate their performance on the \textit{general} summarization. It is important to note that our approach was not re-trained, and it was merely tested on the generic summarization task (1 task per video). Additionally, we allow SubMod to select exactly as many shots as the oracle summary, making it a strong baseline to compare with. For Quasi, we select the reconstruction threshold in a way that the generated summary is roughly the same size as oracle summary. The results are shown in Table~\ref{table:glob}.

\subsection{A Nice Behavior of Our Evaluation Metric}
\vspace{-5pt}
Our evaluation method for video summarization is mainly motivated by Yeung et al.~\cite{yeung2014videoset}. Particularly, we share the same opinion that the evaluation should focus on the semantic information which humans can perceive, rather than the low-level visual features or temporal overlaps. However, the captions used in \cite{yeung2014videoset} are diverse, making the ROUGE-SU4 evaluation unstable and poorly correlated with human judgments~\cite{chen2015microsoft}, and often missing subtle details (cf.\ Figure~\ref{fig:captionvstags} for some examples). 

We rectify those caveats by instead collecting dense concept annotations. Figure~\ref{fig:captionvstags} exhibits a few video shots where the concepts we collected provide a better coverage than the captions about the semantics in the shots. Moreover, we conveniently define an evaluation metric based on the IOU similarity function between any two shots (cf.\ Section~\ref{subsec:tag}) thanks to the concept annotations. 

Our evaluation metric has some nice behaviors. If we randomly remove some video shots from the user summaries and compare the corrupted summaries with the original ones, an accuracy-like metric should give rise to linearly decreasing values. This is indeed what happens to our recall as shown in Figure~\ref{fig:Del}. In contrast, the ROUGE-SU4 recall, taking as input the shot captions, exhibits some nonlinearality. More results on randomly replacing some shots in the user summaries are included in the Suppl. Materials.

%As mentioned in ~\ref{subsec:eval}, summaries are better evaluated when their semantic information are compared. ~\cite{yeung2014videoset} suggests to use ROUGE-SU4, a metric that is used in automatic document summary evaluation. We argue there is more in images than in captions(cf. Figure~\ref{fig:captionvstags}), and proposed to evaluate summaries based with binary indicator vectors where each dimension shows presence/absence of a concept in the shot. To this end, we designed a set of experiments that measures how precision, recall, and F-1 score of ROUGE-SU4 and our novel metric change when randomly removing/replacing elements in a groundtruth summary. This simulation is devised to show the performance drop in the measures as we introduce noise into the summary. To achieve this, we propose to remove/replace 10 up to 70 percent of the user summaries and investigate the trend of deterioration when comparing the corrupted summary against the original user summary. As illustrated in Figure~\ref{fig:Del}, ROUGE-SU4 suffers from a major drop in recall (drops to less than 0.1 when 30 percent of the summary is removed), showing early saturation. On the other hand, our metric, shows an ideal drop in recall. (For the plots of performance deterioration when replacing elements in the summary, please refer to Supp. Material.)

\vspace{-5pt}

%% file: conclude.tex
% !tex root = 1990.tex

\section{Conclusion}
\label{sec:conc}
\vspace{-5pt}
%In this work, by providing query-focused and global summaries for the video in UT Egocentric, we introduced a new benchmark for video summarization that facilitates future works in the field, especially approaches with higher levels of supervision and user preference. In addition, we proposed a novel metric that better incorporates conceptual information in the video into evaluation, hence resulting in a better and fair comparison between system-generated summaries. Finally, we proposed a novel framework for query-focused video summarization, taking advantage of memory-networks that have been proven effective in question-answering tasks.

In this work, our central theme is to study the \textit{subjectiveness} in video summarization. We have analyzed the key challenges caused the subjectiveness and proposed some solutions. In particular, we compiled a dataset that is densely annotated with a comprehensive set of  concepts and designed a novel evaluation metric that benefits from the collected annotations. We also devised a new approach to generating personalized summaries by taking user queries into account. We employed memory networks and determinantal point processes in our summarizer, so that our model leverages their attention schemes and diversity modeling capabilities, respectively. Extensive experiments verify the effectiveness of our approach and reveals some nice behaviors of our evaluation metric.
\vspace{-10pt}
\paragraph{Acknowledgements.}
This work is supported by NSF IIS \#1566511, a gift from Adobe Systems, and a GPU from NVIDIA. We thank Fei Sha, the anonymous reviewers and area chairs, especially R2, for their insightful suggestions.